\pdfoutput=1

\documentclass[11pt]{article}

\usepackage[]{emnlp2021}

\usepackage{times}
\usepackage{latexsym}

\usepackage[T1]{fontenc}

\usepackage[utf8]{inputenc}

\usepackage{microtype}

\usepackage{graphics}
\usepackage{booktabs}
\usepackage{makecell}
\renewcommand\theadfont{\bfseries}
\usepackage{cellspace}
\usepackage{color}
\usepackage{enumitem}
\usepackage{amsmath}
\usepackage{amssymb}
\usepackage{breqn}
\usepackage{hyperref}
\usepackage{siunitx}
\usepackage{dcolumn}
\usepackage{subfigure}
\usepackage{tikz}
\usetikzlibrary{positioning} 
\usepackage{xcolor} 
\usepackage{pgfplots}
\usepackage{tabularx}
\usepackage{array}
\usepackage{amsmath}
\usepackage{setspace}
\usepackage{multirow}
\usepackage{makecell}
\usepackage{colortbl}
\usepackage{hhline}
\title{Learning Implicit Sentiment in Aspect-based Sentiment Analysis\\with Supervised Contrastive Pre-Training}

\author{Zhengyan Li$^1$, Yicheng Zou$^1$, Chong Zhang$^1$, Qi Zhang$^1$\thanks{$^*$  Corresponding author.} \ and Zhongyu Wei$^2$\\
$^1$Shanghai Key Laboratory of Intelligent Information Processing, \\
School of Computer Science, Fudan University, Shanghai, China \\
$^2$School of Data Science, Fudan University \\ 
\texttt{\{lizy19,yczou18,chongzhang20,qz,zywei\}@fudan.edu.cn}}

\begin{document}
\maketitle
\begin{abstract}

Aspect-based sentiment analysis aims to identify the sentiment polarity of a specific aspect in product reviews. We notice that about 30\% of reviews do not contain obvious opinion words, but still convey clear human-aware sentiment orientation, which is known as implicit sentiment. However, recent neural network-based approaches paid little attention to implicit sentiment entailed in the reviews. To overcome this issue, we adopt Supervised Contrastive Pre-training on large-scale sentiment-annotated corpora retrieved from in-domain language resources. By aligning the representation of implicit sentiment expressions to those with the same sentiment label, the pre-training process leads to better capture of both implicit and explicit sentiment orientation towards aspects in reviews. Experimental results show that our method achieves state-of-the-art performance on SemEval2014 benchmarks, and comprehensive analysis validates its effectiveness on learning implicit sentiment.

\end{abstract}

\section{Introduction}
\label{sec:intro}

Aspect-level sentiment analysis (ABSA) is a fine-grained variant aiming to identify the sentiment polarity of one or more mentioned aspects in product reviews. 
Recent studies tackle the task by either employing attention mechanisms \citep{wang-etal-2016-attention, ijcai2017-568} or incorporating syntax-aware graph structures \citep{he-etal-2018-effective, tang-etal-2020-dependency, zhang-etal-2019-aspect, sun-etal-2019-aspect, wang-etal-2020-relational}.
Both methodologies aim to capture the corresponding sentiment expression towards a particular aspect, which is usually an opinion word that explicitly expresses sentiment polarity. 
For instance, given the review on a restaurant \emph{``Great food but the service is dreadful''}, current models attempt to find \emph{``great''} for aspect \emph{``food''} to determine the positive sentiment polarity towards it. 

\newcommand\positive[1]{\colorbox{red!15}{#1}}
\newcommand\negative[1]{\colorbox{blue!15}{#1}}
\newcommand\background[1]{\colorbox{red!15}{#1}}
\renewcommand\theadfont{\bfseries}
\begin{table}
\centering
\setlength\tabcolsep{-2pt}
\scalebox{0.75}{
\begin{tabular}{>{\hspace{0.3pc}}l<{\hspace{0.3pc}}}
\bottomrule
\\ [-1em]
\makecell[c]{\textbf{\quad \quad \ \ Reviews contain implicit sentiment}}  \\
\\ [-1.2em]
\bottomrule
\\ [-1em]
The \textbf{waiter} poured water on my hand and walked away \\
The \textbf{bartender} continued to pour champagne from his reserve \\
\midrule
10 hours of \textbf{battery life} ...  \\
The \textbf{battery life} is probably an hour \\
\\ [-1.3em]
\toprule
\end{tabular}
}
\caption{\label{tab:examples} Examples of reviews contain implicit sentiment where aspects are marked to bold. In the above examples, \emph{``pour''} expresses opposite emotions in different contexts. In the below examples, people determine the sentiment orientations towards \emph{``battery''} by referring to a common lifetime.}
\end{table}

However, implicit sentiment expressions widely exist in the recognition of aspect-based sentiment. 
Implicit sentiment expressions indicate sentiment expressions that contain no polarity markers but still convey clear human-aware sentiment polarity in context \citep{russo-etal-2015-semeval}.
As illustrated in Table \ref{tab:examples}, the comment \emph{``The waiter poured water on my hand and walked away''} towards aspect \emph{``waiter''} contains no opinion words, but can be clearly interpreted to be negative.
According to Table \ref{tab:dataset} (as seen in Section \ref{sec:dataset}), 27.47\% and 30.09\% of reviews contain implicit sentiment among Restaurant and Laptop datasets. 
However, most of the previous methods generally pay little attention on modeling implicit sentiment expressions.
This motivates us to better solve the task of ABSA by capturing implicit sentiment in an advanced way.



To equip current models with the ability to capture implicit sentiment, inadequate ABSA datasets are the main challenge.
With only a few thousand labeled data, models could hardly recognize comprehensive patterns of sentiment expressions, and are unable to capture enough commonsense knowledge, which is required in sentiment identification. 
It reveals that external sentiment knowledge should be introduced to solve the problem.


Therefore, we adopt Supervised ContrAstive Pre-Training(SCAPT) on external large-scale sentiment-annotated corpora to learn sentiment knowledge. 
Supervised contrastive learning gives an aligned representation of sentiment expressions with the same sentiment label. 
In embedding space, explicit and implicit sentiment expressions with the same sentiment orientation are pulled together, and those with different sentiment labels are pushed apart. 
Considering the sentiment annotations of retrieved corpora are noisy, supervised contrastive learning enhances noise immunity of the pre-training process. 
Also, SCAPT contains review reconstruction and masked aspect predication objectives. The former requires representation encoding review context besides sentiment polarity, and the latter adds the model’s ability to capture the sentiment target. Overall, the pre-training process captures both implicit and explicit sentiment orientation towards aspects in reviews.

Experimental evaluations conducted on SemEval-2014 \citep{pontiki-etal-2014-semeval} and MAMS \citep{jiang-etal-2019-challenge} datasets show that proposed SCAPT outperforms baseline models by a large margin. The results on partitioned datasets demonstrate the effectiveness of both implicit sentiment expression and explicit sentiment expression. Moreover, the ablation study verifies that SCAPT efficiently learns implicit sentiment expression on the external noisy corpora. Codes and datasets are publicly available\footnote{\href{https://github.com/Tribleave/SCAPT-ABSA}{https://github.com/Tribleave/SCAPT-ABSA}}.

The contributions of this work include:

\begin{itemize}[leftmargin=*,noitemsep,topsep=0pt]
    \item We reveal that ABSA was only marginally tackled by previous studies since they paid little attention to implicit sentiment. 
    \item We propose Supervised Contrastive Pre-training to learn sentiment knowledge from large-scale sentiment-annotated corpora.
    \item Experimental results show that our proposed model achieves state-of-the-art performance, and is effective to learn implicit sentiment.

\end{itemize}

\begin{figure*}[htbp]
  \centering
  \includegraphics[width=0.85\textwidth]{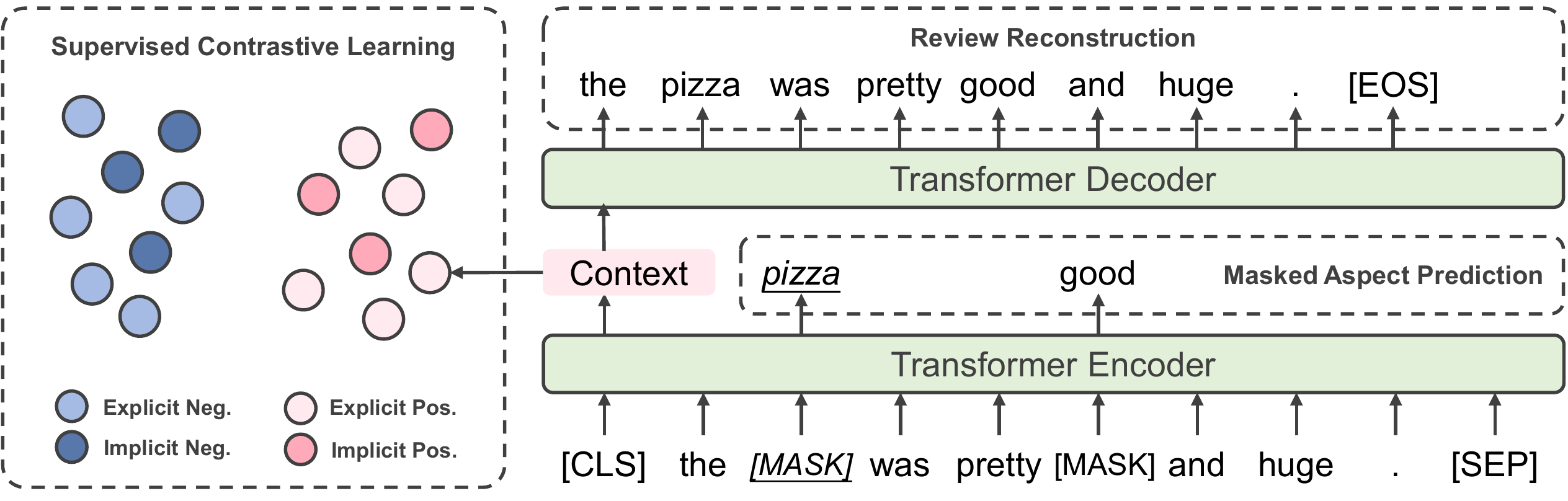}
  \caption{\label{fig: fig1} 
  An overview of SCAPT on ABSA. SCAPT consists of three objectives, in which Supervised Contrastive Learning aligns the representations with the same sentiment label. 
}
\end{figure*}

\section{Implicit Sentiment}
\label{sec:IS}
As sentiment that can only be inferred within the context of reviews, many researches address the presence of implicit sentiment in sentiment analysis. 
\citet{toprak-etal-2010-sentence, russo-etal-2015-semeval} proposed similar terminologies (as \textit{implicit polarity} or \textit{polar facts}), and provided corpora containing implicit sentiment.
\citet{deng-wiebe-2014-sentiment} detected implicit sentiment via inference over explicit sentiment expressions and so-called goodFor/badFor events. \citet{choi-wiebe-2014-effectwordnet} used +/-EffectWordNet lexicon to identify implicit sentiment, by assuming sentiment expressions are often related to states and events which have positive/negative/null effects on entities. 

To investigate the ubiquitous of implicit sentiment in ABSA, we split SemEval-2014 Restaurant and Laptop benchmarks into Explicit Sentiment Expression (ESE) slice and Implicit Sentiment Expression (ISE) slice, based on the presence of opinion words. \citet{fan-etal-2019-target} have annotated opinion words for target aspects on SemEval benchmarks. 
We notice that the provided datasets do not keep the original order and have some differences in texts. 
Thus, we first match the annotations to the original datasets, and then manually pick the reviews including opinion words towards the aspect from the remaining part. As results shown in Table \ref{tab:dataset} (as seen in Section \ref{sec:dataset}), 27.47\% and 30.09\% of reviews are divide into ISE part among Restaurant and Laptop, revealing that implicit sentiment exists widely in ABSA and is worthy to be explored.

\section{Methodology}

In this section, we introduce the pre-training and fine-tuning scheme of our models. In pre-training, we introduce Supervised ContrAstive Pre-Training (SCAPT) for ABSA, which learns the polarity of sentiment expressions by leveraging retrieved review corpus. In fine-tuning, aspect-aware fine-tuning is adopted to enhance the ability of models on aspect-based sentiment identification. 

\subsection{Supervised Contrastive Pre-training}

Three objectives are included in SCAPT: supervised contrastive learning, masked aspect prediction, and review reconstruction. The details of SCAPT's procedure are shown in Figure \ref{fig: fig1}. 

\paragraph{Transformer Encoder Backbone}
The pre-training scheme is built on Transformer encoder \citep{vaswani2017attention}. We denote the retrieved review corpus used in SCAPT as $D = \{\mathbf{x}_1, \mathbf{x}_2, \ldots, \mathbf{x}_n \}$ including $n$ sentences. The $i$-th sentence $\mathbf{x}_i$ is labeled with $y_i$.  For each input sentence $\mathbf{x}_i$, following \citet{devlin-etal-2019-bert}, we format the input sentence as $I_i = \mathrm{[CLS]} + \mathbf{x}_i + \mathrm{[SEP]}$ to feed into the model. 
The output vector of $\mathrm{[CLS]}$ token encodes the sentence representation $\bar{h}_i$: 
\begin{equation}
    \bar{h}_i, \dots = \mathrm{TransEnc}(I_i)
\end{equation}

\paragraph{Supervised Contrastive Learning} 
Inspired by \citet{khosla2020supervised}, we adopt supervised contrastive learning objective in SCAPT to align the representation of explicit and implicit sentiment expressions with the same emotion. 
Supervised contrastive learning encourages the model to capture the entailed sentiment orientation in context and incorporate it in sentiment representation. 

Specifically, for $\left( \mathbf{x}_i, y_i \right)$ within a batch $B$, we first extract sentiment representation $s_i = \mathbf{W_s} \bar{h}_i$ from sentence representation $\bar{h}_i$ of $\mathbf{x}_i$. $\mathbf{W_s}$ could be seen as a trainable sentiment perceptron for sentences. The supervised contrastive loss on the batch $B$ is defined as:
\begin{gather}
    P^{sup}_B(i, c) = \frac{\exp{\left(sim(s_i, s_c) / \tau \right)}}{
    {\sum_{\substack{b \in B, b \neq i}}}{ \exp{\left(sim(s_i, s_b) / \tau \right)}}}\\
    \mathcal{L}^{sup}_B = \sum_{i \in B}{ - \log \frac{1}{C_i} \sum_{\substack{y_i = y_c, c \neq i}}{P^{sup}_{B}(i, c)}}
    \label{func:pre-loss-1}
\end{gather}
Here, $P^{sup}_B(i, c)$ indicates the likelihood that $s_c$ is most similar to $s_i$ and $\tau$ is the temperature of softmax. Here we simply use 
$sim(s_i, s_c) = s_i \cdot s_c$ for similarity metric. Supervised contrastive loss $\mathcal{L}^{sup}_B$ is calculated for every sentence $s_i$ among $B$, where $C_i = \left|\{c | y_c = y_i, c \neq i \}\right|$ is the number of samples in the same category $y_i$ in $B$. Notably, we do not directly use sentence representation in the supervised contrastive pre-training process. Instead, we use sentiment representation to make full use of document-level labeled corpora in mining the inherent sentiment perception. 

\paragraph{Review Reconstruction}
Motivated by the power of denoising auto-encoder \citep{vincent2008extracting} and its success in pre-training models \citep{lewis-etal-2020-bart}, we further propose review reconstruction task to enhance the sentence representation on context semantic modeling. 
With solely pre-trained on the supervised contrastive learning task which only focuses on sentiment regularization, the essential semantic information is not completely preserved in the sentence representations. Thus, we additionally employ review reconstruction in SCAPT to capture comprehensive context information in sentence representations.

Generally, this objective reconstructs the whole sentence $\mathbf{x}_i$ with the sentence representation $\bar{h}_i$. 
After encoding $\mathbf{x}_i$ to the sentence representation $\bar{h}_i$, the latter is fed to Transformer decoder for auto-regressive generation:
\begin{equation}
    P^{rec}(\Tilde{\mathbf{x}}_i) = \mathrm{TransDec}(\bar{h}_i)
\end{equation}
$\Tilde{\mathbf{x}}_i$ is the recovered sentence. $\bar{h}_i$ acts as a beginning-of-sentence input embedding in the decoding process to control the whole generation. 
We use the original sentence $\mathbf{x}_i$ without masking as the gold reference of review reconstruction objective:
\begin{equation}
    \mathcal{L}_i^{rec} = - \log P^{rec}(\Tilde{\mathbf{x}}_i)
    \label{func:pre-loss-3}
\end{equation}

\paragraph{Masked Aspect Prediction} 
In masked aspect prediction, the model learns to predict the masked aspect from a corrupted version for each review. 
The masking strategy of input reviews consists of following two steps:
\begin{enumerate}[leftmargin=*,noitemsep,topsep=0pt]
    \item Aspect Span Masking. Since all inputs are from our retrieved corpora, we ensure that each review contains at least one aspect. For each input, the tokens of aspect spans are replaced with $\mathrm{[MASK]}$ with 80\% probability, or replaced with a random token with 10\% probability, otherwise kept unchanged. Aspect span masking provides a better capture of aspect words.
    \item Random Masking. After aspect span masking, if the proportion of masked tokens is less than 15\%, we randomly mask extra tokens from the rest ones to reach the proportion. 
\end{enumerate}


We denote the input token of $\mathrm{[MASK]}$ as $w_{\textit{MASK}}$.
For each masked input token at $k$-th position, its contextualized hidden representation $h_{ik}$ is fed into a softmax layer to predict the original word: 
\begin{equation}
    P^{map}(k) = \mathrm{softmax}(\mathbf{W}_o h_{ik})
\end{equation}
Specific to the above equation, $h_{ik}$ is the output of Transformer encoder at $k$-th position, $\mathbf{W}_o$ is a trainable parameter matrix, and $P^{map}(k)$ indicates the predict probability of the original word at $k$-th position.
The masked aspect prediction loss is an accumulation of log-likelihood on predictions of each masked position: 
\begin{equation}
    \mathcal{L}^{map}_i = \sum_{\substack{x_{ik} = w_{\textit{MASK}}}}{-\log{P^{map}(k)}}
    \label{func:pre-loss-2}
\end{equation}

Different from MLM \citep{devlin-etal-2019-bert} or sentiment masking \citep{tian-etal-2020-skep}, masked aspect prediction focuses more on modeling aspect-related context information in aspect-based representations, which complements the other pre-training objectives and purposefully benefits our fine-tuning scheme. 


\paragraph{Joint Training} 
The three losses mentioned above are combined and jointly trained in SCAPT. For the overall pre-training loss $\mathcal{L}^{pre}_B$ on batch $B$, the review reconstruction loss and masked aspect prediction loss are counted on each example $b \in B$, and $\alpha$ and $\beta$ are coefficients to balance the objectives:

\begin{equation}
    \mathcal{L}^{pre}_B = \mathcal{L}^{sup}_B + 
                          \alpha \sum_{b\in B}{\mathcal{L}^{rev}_b} +
                          \beta  \sum_{b\in B}{\mathcal{L}^{map}_b}
\end{equation}


\begin{figure}[tbp]
  \centering
  \includegraphics[width=0.4\textwidth]{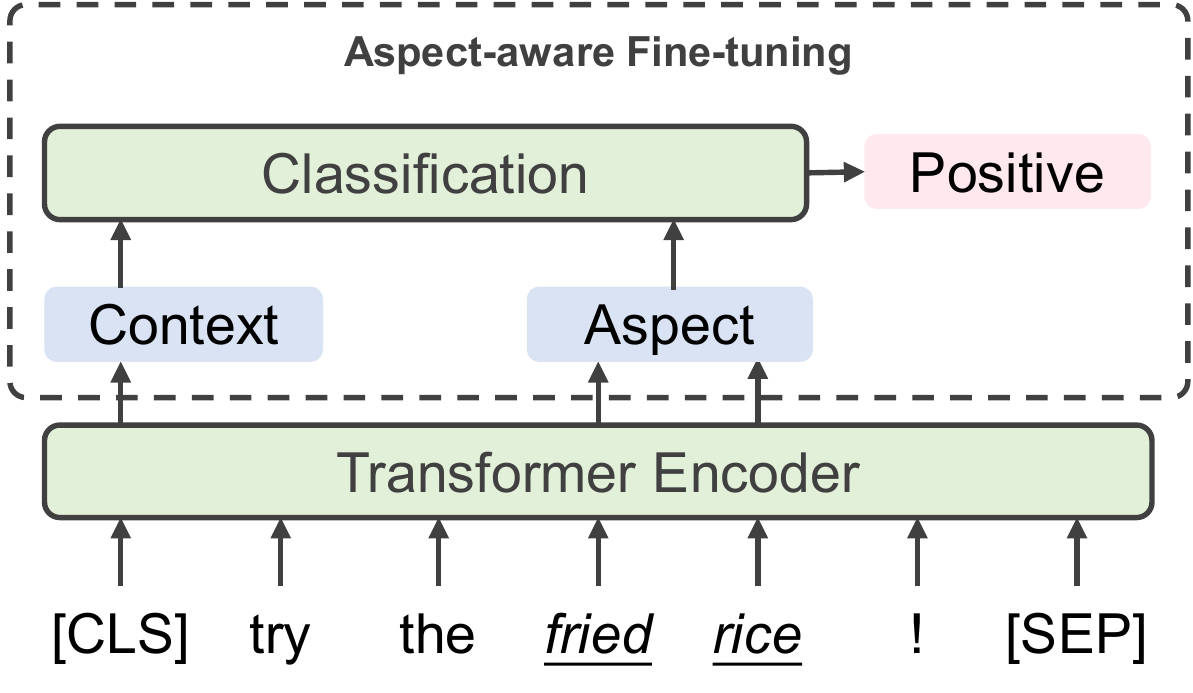}
  \caption{\label{fig: fig2} 
  Aspect-aware fine-tuning on Transformer encoder based models. Sentiment representation and aspect-based representation are taken into account in sentiment classification. 
}
\end{figure}

\subsection{Aspect-Aware Fine-tuning} 
Our proposed models are fine-tuned on ABSA benchmarks by aspect-aware fine-tuning, to fully leverage their ability of sentiment identification. They also learn to capture aspect-related sentiment information during fine-tuning. 

Specifically, given a sentence $\mathbf{x}^{ab} = \{w_{1}, \dots, w_{a}, \dots w_{n} \}$ in ABSA dataset $D^{ab}$, and $w_{a}$ is one of the aspects occurring in $\mathbf{x}^{ab}$. In fine-tuning, models predict aspect-level sentiment orientation $y_a^{ab}$ according to aspect-based representation $\bar{h}^{ab}_a$ and sentiment representation $s^{ab}$. 

\paragraph{Aspect-based Representation}
The research \citep{ethayarajh-2019-contextual} on pre-trained contextualized word representation has demonstrated that it can capture context information related to the word. 
Thus, in spite of using laborious methods to embed the aspect information, we extract aspect-based representation $\bar{h}^{ab}_a$ by collecting final hidden states that correspond to $w_{a}$. 
In fine-tuning, $\bar{h}^{ab}_a$ would focus on aspect-related words in context, which we believe would enhance the perception of aspect-specific opinion words and bring the model with a good view of explicit sentiment. 
Specifically, let $\mathcal{I}_a$ be the token index in aspect $x_{a}$, we average the hidden state $h_i$ for all $i \in \mathcal{I}_a$ to acquire aspect-based representation:
\begin{equation}
    \bar{h}^{ab}_a = \mathrm{AveragePooling}_{i \in \mathcal{I}_a}(h_i)
\end{equation}
Notably, when processing multiple aspects $w_{a1}, w_{a2}, \dots$ in sentence $\mathbf{x}^{ab}$, we extract aspect-based representation $\bar{h}^{ab}_{a1}, \bar{h}^{ab}_{a2}, \dots$ in a single run, while previous methods embed aspect and encoder whole input for each aspect one-by-one.

\paragraph{Representation Combination}
For sentiment classification, aspect-based representation and sentiment representation are considered jointly to predict aspect-level sentiment polarity. In that case, fine-tuned model builds the perception of both word-occurrence-related explicit sentiment and semantic-related implicit sentiment. We use the same sentiment perceptron $\mathbf{W}_s$ in pre-training to extract sentiment representation $s^{ab}$ from sentence representation.
Then sentiment representation $s^{ab}$ and aspect-based representation $\bar{h}^{ab}_a$ are concatenated for predicting aspect-level sentiment polarity:
\begin{equation}
    {y}_a^{ab} = \mathrm{softmax}{\left( \mathbf{W}_a[s^{ab}; \bar{h}^{ab}_a] \right)}
\end{equation}
${y}_a^{ab}$ is the prediction on aspect $x_a$ and $\mathbf{W}_a$ is trainable parameter matrix. Lastly, our fine-tuning objective is cross-entropy loss for prediction task $\mathcal{L}^{ab}=-\sum_{\mathbf{x}^{ab}\in D^{ab}}{\log{y_{a}^{ab}}}$. 


\newcommand{\tabincell}[2]{\begin{tabular}{@{}#1@{}}#2\end{tabular}}  
\newcolumntype{C}[1]{m{#1}<{\centering}}

\section{Experimental Settings}



\label{sec:dataset}
\paragraph{ABSA Datasets}
Our experiments are mainly conducted on two benchmarks, Laptop and Restaurant review from SemEval 2014 task 4 \citep{pontiki-etal-2014-semeval}. We use ESE and ISE slices of their test parts to evaluate model performance on explicit and implicit sentiment respectively. The process to build these slices is detailed in Section \ref{sec:IS}. Furthermore, we also use a more challenging dataset, Multi-Aspect Multi-Sentiment (MAMS) \citep{jiang-etal-2019-challenge}, which shares the same domain to SemEval2014 Restaurant. All these datasets involve three sentiment categories which are positive, neutral, and negative. The details of these ABSA datasets can be found in Table \ref{tab:dataset}.

\paragraph{Retrieved External Corpora}
We retrieve large-scale sentiment-annotated corpora from document-level labeled data for pre-training. Specifically, we first extract five-stars-rated/one-star-rated reviews from the Yelp\footnote{\url{https://www.yelp.com/dataset}} and Amazon Review \citep{he2016ups} datasets, and label them as positive/negative. 
Such a procedure can mitigate the noise in the 5-way rated document-level sentiment language source.
Then we preserve reviews within the topic of restaurant/laptop to make sure that pre-train corpora and ABSA datasets are in the same domain. Later, we split these document-level reviews into sentences and preserve sentences containing the same aspect term as those mentioned in ABSA training sets. The sentiment label of each sentence is determined by the label of its original review. After the retrieving process, we finally acquire about 1.56/0.51 million sentence-level reviews from Yelp/Amazon that are noisy-labeled as positive/negative. After manually checking a small portion of both corpora, we confirm that both implicit and explicit sentiment expressions are available.
We pre-train our models on the retrieved corpus that shares the same domain with the downstream ABSA task. Specifically, we adopt Yelp when dealing with Restaurant or MAMS, and Amazon for Laptop. 

\renewcommand\tabcolsep{3.5pt}

\begin{table}[t]
\scriptsize
\centering
\begin{spacing}{1.19}
\begin{tabular}{l|cccccl}
\bottomrule
Dataset & Positive & Neutral & Negative & Total & \makecell[c]{Implicit \\ Sentiment}\ \% \\
\hline
Restaurant-train & 2164 & 805 & 633 & 3602 & 28.59 \ \ \ \ \  \\
Restaurant-test & 728 & 196 & 196 & 1120 & 23.84\ \ \ \ \  \\
Restaurant & 2892 & 1001 & 829 & 4722 & 27.47\ \ \ \ \  \\
Laptop-train & 987 & 866 & 460 & 2313 & 30.87\ \ \ \ \  \\
Laptop-test & 341 & 128 & 169 & 638 & 27.27\ \ \ \ \  \\
Laptop & 1328 & 994 & 629 & 2951 & 30.09\ \ \ \ \  \\
MAMS & 4183 & 6253 & 3418 & 13854 & -\ \ \ \ \  \\ 
\hline
YELP & 1.17M & - & 0.39M & 1.56M & -\ \ \ \ \  \\
Amazon & 0.38M & - & 0.13M & 0.51M & -\ \ \ \ \  \\
\toprule
\end{tabular}
\end{spacing}
\caption{Statistics on three datasets of ABSA and two external corpus for SCAPT.}
\label{tab:dataset}
\end{table}

\paragraph{Models with SCAPT}
We apply SCAPT to Transformer encoder and BERT, and these models are fine-tuned by aspect-aware fine-tuning. The models are so-called TransEncAsp+SCAPT and BERTAsp+SCAPT respectively. We use a 300-dimensional randomly initialized Transformer encoder with 6 layers and 6 heads and BERT-base-uncased as the basis. The pre-training for Transformer encoder and BERT takes 80 and 8 epochs respectively. We adopt Adam \citep{DBLP:journals/corr/KingmaB14} with warm-up to optimize our models with learning rate $1\mathrm{e}{-3}$ for Transformer encoder and $5\mathrm{e}{-5}$ for BERT. The pre-trained models are fine-tuned by aspect-aware fine-tuning with $5\mathrm{e}{-5}$ learning rate. The hyper-parameters are set as $\alpha = \beta = 1$ for combining objectives in SCAPT, and $\tau = 0.07$ in supervised contrastive learning. 

\renewcommand\tabcolsep{5pt}
\newcolumntype{W}{C{0.05\textwidth}}
\definecolor{mygray}{gray}{.9}
\begin{table*}[t]
\small
\centering
\begin{spacing}{1.19}
\begin{tabular}{c|l|WWWW|WWWW}
\bottomrule
\multicolumn{2}{c|}{\multirow{2}{*}{\textbf{Method}}} & \multicolumn{4}{c|}{\textbf{Restaurant}} & \multicolumn{4}{c}{\textbf{Laptop}} \\
\cline{3-10}
\multicolumn{2}{c|}{} & Acc. & F1 & ESE & ISE & Acc. & F1 & ESE & ISE \\
\hline
\multirow{4}{*}{\makecell[c]{Attention}} 
 & ATAE-LSTM \citep{wang-etal-2016-recursive} & \text{\,76.90*} & \text{\,62.64*} & 84.16 & 53.71 & \text{\,65.37*} & \text{\,62.92*} & 75.69 & 37.86 \\
 & IAN \citep{ijcai2017-568} & \text{\,76.88*} & \text{\,67.71*} & 86.52 & 46.07 & \text{\,67.24*} & \text{\,63.72*} & 75.86 & 44.25 \\
 & RAM \citep{chen-etal-2017-recurrent-attention} & 80.23 & 70.80 & 85.11 & 55.81 & 74.49 & 71.35 & 75.86 & 44.25 \\
 & MGAN \citep{fan-etal-2018-multi} & 81.25 & 71.94 & 85.18 & 60.04 & 75.39 & 72.47 & 76.16 & 56.31 \\
\hline
\multirow{4}{*}{GNN} 
 & ASGCN \citep{zhang-etal-2019-aspect} & 80.77 & 72.02 & 84.29 & 62.91 & 75.55 & 71.05 & 75.46 & 57.77 \\
 & BiGCN \citep{zhang-qian-2020-convolution} & 81.97 & 73.48 & 87.19 & 59.05 & 74.59 & 71.84 & 79.53 & 62.64 \\
 & CDT \citep{sun-etal-2019-aspect} & 82.30 & 74.02 & 88.79 & 65.87 & 77.19 & 72.99 & 77.53 & 68.90 \\
 & RGAT \citep{wang-etal-2020-relational} & 83.30 & 76.08 & 89.45 & 61.05 & 77.42 & 73.76 & 80.17 & 65.52 \\
\hline
\multirow{5}{*}{\makecell[c]{Knowledge \\ Enhanced}} 
 & TransCap \citep{chen-qian-2019-transfer} & 79.55 & 71.41 & 86.52 & 59.93 & 73.87 & 70.10 & 77.16 & 60.34 \\
 & BERT-SPC \citep{devlin-etal-2019-bert} & \text{\,83.57*} & \text{\,77.16*} & 89.21 & 65.54 & \text{\,78.22*} & \text{\,73.45*} & 81.47 & 69.54 \\
 & CapsNet+BERT \citep{jiang-etal-2019-challenge} & \text{\,85.09*} & \text{\,77.75*} & 91.68 & 64.04 & \text{\,78.21*} & \text{\,73.34*} & 82.33 & 67.24  \\
 & BERT-PT \citep{xu-etal-2019-bert} & 84.95 & 76.96 & 92.15 & 64.79 & 78.07 & 75.08 & 81.47 & 71.27 \\
 & BERT-ADA \citep{rietzler-etal-2020-adapt} & 87.14 & 80.05 & 94.14 & 65.92 & 78.96 & 74.18 & 82.76 & 70.11 \\
 & R-GAT+BERT \citep{wang-etal-2020-relational} & 86.60 & 81.35 & 92.73 & 67.79 & 78.21 & 74.07 & 82.44 & 72.99 \\
\hline
\multirow{5}{*}{\textbf{Ours}} 
 & TransEncAsp & 77.10 & 57.92 & 86.97 & 48.96 & 65.83 & 59.53 & 74.31 & 43.20 \\
 & BERTAsp & 85.80 & 78.95 & 92.73 & 63.67 & 78.53 & 74.07 & 82.33 & 68.39 \\
 & BERTAsp+CEPT & 87.50 & 82.07 & 93.67 & 67.79 & 81.66 & 78.38 & 83.84 & 75.86 \\
 \hhline{~|---------}
 & \cellcolor{mygray}TransEncAsp+SCAPT & \cellcolor{mygray}83.39 & \cellcolor{mygray}74.53 & \cellcolor{mygray}88.04 & \cellcolor{mygray}68.55 & \cellcolor{mygray}77.17 & \cellcolor{mygray}73.23 & \cellcolor{mygray}78.70 & \cellcolor{mygray}72.82 \\
 & \cellcolor{mygray}BERTAsp+SCAPT & \cellcolor{mygray}\textbf{89.11} & \cellcolor{mygray}\textbf{83.79} & \cellcolor{mygray}\textbf{94.37} & \cellcolor{mygray}\textbf{72.28} & \cellcolor{mygray}\textbf{82.76} & \cellcolor{mygray}\textbf{79.15} & \cellcolor{mygray}\textbf{84.70} & \cellcolor{mygray}\textbf{77.59} \\
\toprule
\end{tabular}
\end{spacing}
\caption{Overall performance of different methods on Restaurant and Laptop. We rerun the code of baselines and report their accuracy on ESE and ISE slices of the two datasets. For the baselines of which the accuracy or F1-score is missing, we also report the accuracy and F1-score of our rerunning version, and these results are marked with *.}
\label{tab:main}
\end{table*}

\paragraph{Baselines}
We compare the proposed models with baselines from different perspectives to comprehensively evaluate the performance of our approach:
\begin{itemize}[leftmargin=*,itemsep=0pt,topsep=0pt]
    \item \textbf{Attention-based models}:
        ATAE-LSTM \citep{wang-etal-2016-attention}, 
        IAN \citep{ijcai2017-568}, 
        RAM \citep{chen-etal-2017-recurrent-attention},
        and MGAN \citep{fan-etal-2018-multi}.
    \item \textbf{Graph neural networks}:
        ASGCN \citep{zhang-etal-2019-aspect}, 
        BiGCN \citep{zhang-qian-2020-convolution}, 
        CDT \citep{sun-etal-2019-aspect},
        and RGAT \citep{wang-etal-2020-relational}.
    \item \textbf{Knowledge-enhanced methods}:
        TransCap \citep{chen-qian-2019-transfer}, 
        BERT-SPC \footnote{BERT-SPC denotes fine-tuning BERT for Sentence Pair Classification, in which the input review is transformed to ``[CLS] + context + [SEP] + aspect + [SEP]'' and fed into BERT for classification \citep{song2019attentional}.} \citep{devlin-etal-2019-bert}, 
        CapsNet+BERT \citep{jiang-etal-2019-challenge}, 
        BERT-PT \citep{xu-etal-2019-bert}, 
        BERT-ADA \citep{rietzler-etal-2020-adapt},
        and R-GAT+BERT \citep{wang-etal-2020-relational}.
\end{itemize}
For better analyze the effect of SCAPT and aspect-aware fine-tuning, we further propose the following variants as baselines:
\begin{itemize}[leftmargin=*,itemsep=0.1pt,topsep=0pt]
    \item \textbf{TransEncAsp}: Directly apply aspect-aware fine-tuning on randomly initialized Transformer encoder without pre-training.
    \item \textbf{BERTAsp}: Directly apply aspect-aware fine-tuning on BERT-base.
    \item \textbf{BERTAsp+CEPT}: Merely replace the supervised contrastive learning loss with cross-entropy loss in SCAPT. Other settings are the same as BERTAsp+SCAPT.
\end{itemize}

\section{Results and Analysis}

This section mainly demonstrates the experiment results. Our model achieves state-of-the-art on three ABSA benchmarks, and we illustrate the representation alignment effect of supervised contrastive learning and the effectiveness of other parts from several perspectives. 
Moreover, we reveal that our model is capable to identify implicit sentiment, and attributes its effectiveness to supervised contrastive learning in SCAPT. 


\subsection{Main Results}

The performance of baselines and our proposed models are shown in Table \ref{tab:main}. Models are evaluated with Accuracy and Macro-F1. According to the results, several observations can be noted. 

\textbf{Our model achieves SOTA performance.} BERTAsp+SCAPT outperforms the current SOTA model by 1.97\%/3.80\% on Restaurant/Laptop. TransEncAsp+SCAPT performs better than most baselines without pre-trained knowledge. Moreover, BERTAsp+SCAPT also achieves the best performance on ESE/ISE slices of the two datasets, revealing the effectiveness of the proposed pre-training scheme.

\textbf{After pre-trained with SCAPT, models improve significantly on ABSA tasks.} Compared with BERTAsp which directly fine-tuned on ABSA datasets, BERTAsp+SCAPT achieves a 3.31\%/4.23\% performance gain on Restaurant/Laptop, which is a convinced proof that acquiring in-domain knowledge with proper adaptive pre-training is still necessary for knowledge-enhanced models, and SCAPT is an effective approach to be adopted. Moreover, TransEncAsp+SCAPT is 6.29\%/11.34\% better that TransEncAsp, illustrating that incorporating sentiment knowledge with SCAPT greatly potentiates ABSA models.

\textbf{SCAPT is good at learning implicit sentiment.} This could be verified from several perspectives. 
First, compared with its performance on ESE, BERTAsp+SCAPT appears to be much better on ISE. Compared with other works, BERTAsp+SCAPT is around 0-2\% better on ESE slices, but surpasses the previous SOTA model by 4.49\%/4.60\% on ISE slices. Therefore, the well performance of BERTAsp+SCAPT mainly contributes to its awareness of implicit sentiment.
Second, TransEncAsp+SCAPT behaves much better than BERTAsp on ISE slices. With only exposing to million-scale pre-training corpus, TransEncAsp+SCAPT is generally worse than BERTAsp on the whole task, but exceeds BERTAsp by 4.88\%/4.43\% on ISE slices. This demonstrates that SCAPT is data-effective on learning implicit sentiment. 
Last, after pre-trained with SCAPT, models attain remarkable performance gain on ISE which is much more significant than ESE. BERTAsp+SCAPT is ~2\% better than BERTAsp on ESE, but outperforms the latter by 8.61\%/9.20\% on ISE. As for Transformer encoder based models, the performance gain on ISE after SCAPT goes beyond 20\%. We conclude that what models have learned in SCAPT is dominantly the perception of implicit sentiment.

\textbf{Aspect-aware fine-tuning serves as a complement to SCAPT.} We find that models with aspect-aware fine-tuning perform better on ESE slices of the datasets. Specifically, BERTAsp performs worse on ISE but better on ESE compared with BERT-SPC, and is therefore evaluated to be better on the two datasets. The better performance of BERTAsp on ESE slices may mainly due to its use of aspect-based representation, which attends to aspect-related context that may contain sentiment orientation. This characteristic of aspect-aware fine-tuning makes it suitable to enhance the recognition of explicit sentiment of models pre-trained with SCAPT.

\newcolumntype{X}{C{0.07\textwidth}}
\begin{table}[t]
\small
\centering
\begin{spacing}{1.19}
\begin{tabular}{l|XX}
\bottomrule
\multirow{2}{*}{\textbf{Method}} & \multicolumn{2}{c}{\textbf{MAMS}} \\ \cline{2-3}
 & Acc. & F1 \\ \hline
ATAE-LSTM & 77.05 & - \\
IAN & 76.60 & - \\
CapsNet & 79.78 & - \\
BERT-SPC & 82.22 & - \\
CapsNet+BERT & 83.39 & - \\
BERTAsp & 84.20 & 83.82 \\
\hline
TransEncAsp+SCAPT & 80.54 & 79.83 \\
BERTAsp+SCAPT & \textbf{85.63} & \textbf{85.24} \\
\toprule
\end{tabular}
\end{spacing}
\caption{Model performance on MAMS.}
\label{tab:mams}
\end{table}

\subsection{Effectiveness on Multi-aspect ABSA}
Table \ref{tab:mams} shows the performance of baselines and our models in MAMS datasets. Though it is challenging to distinguish the sentiment polarities of multiple aspects in a single sentence, the result shows TransEncAsp+SCAPT outperforms baselines that lack external sentiment knowledge, and BERTAsp+SCAPT achieves state-of-the-art in the multi-aspect scenario. The efficiency of our models can attribute to both SCAPT and aspect-aware fine-tuning since they enhance the learning of implicit and explicit sentiment respectively. 
Besides, BERTAsp performs much more better than BERT-SPC in MAMS than in Restaurant/Laptop. We suppose the exceeding performance of BERTAsp credits to its modeling of contextual information in aspect-based representation, which is more important in multi-aspect ABSA. 

\begin{figure*}[tbp]
  \hspace{-0\textwidth}
  \centering
  \subfigure[width=0.37\textwidth][BERTAsp]{
  \includegraphics[width=0.37\textwidth]{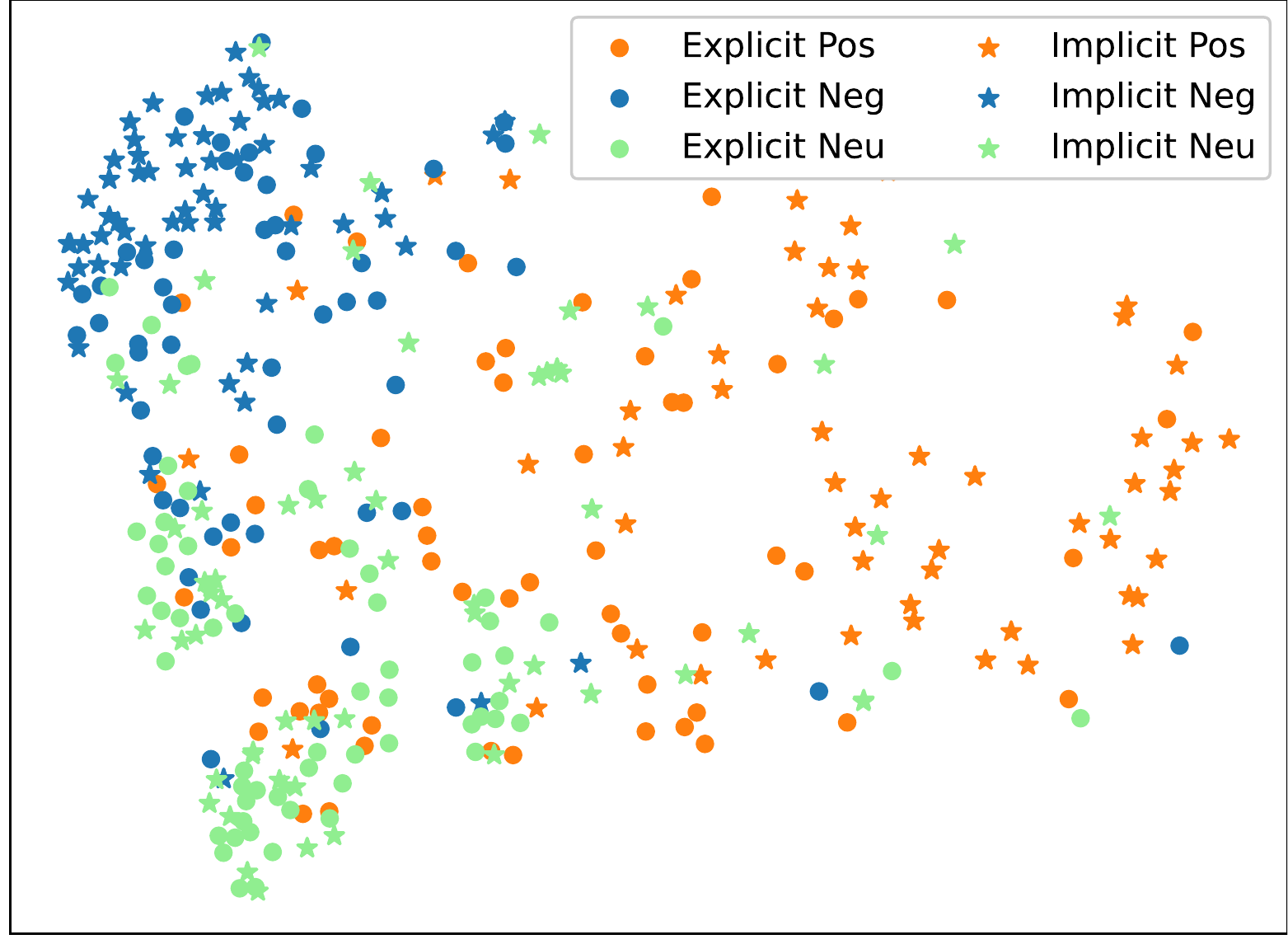}
  }
  \subfigure[width=0.37\textwidth][BERTAsp+SCAPT]{
  \includegraphics[width=0.37\textwidth]{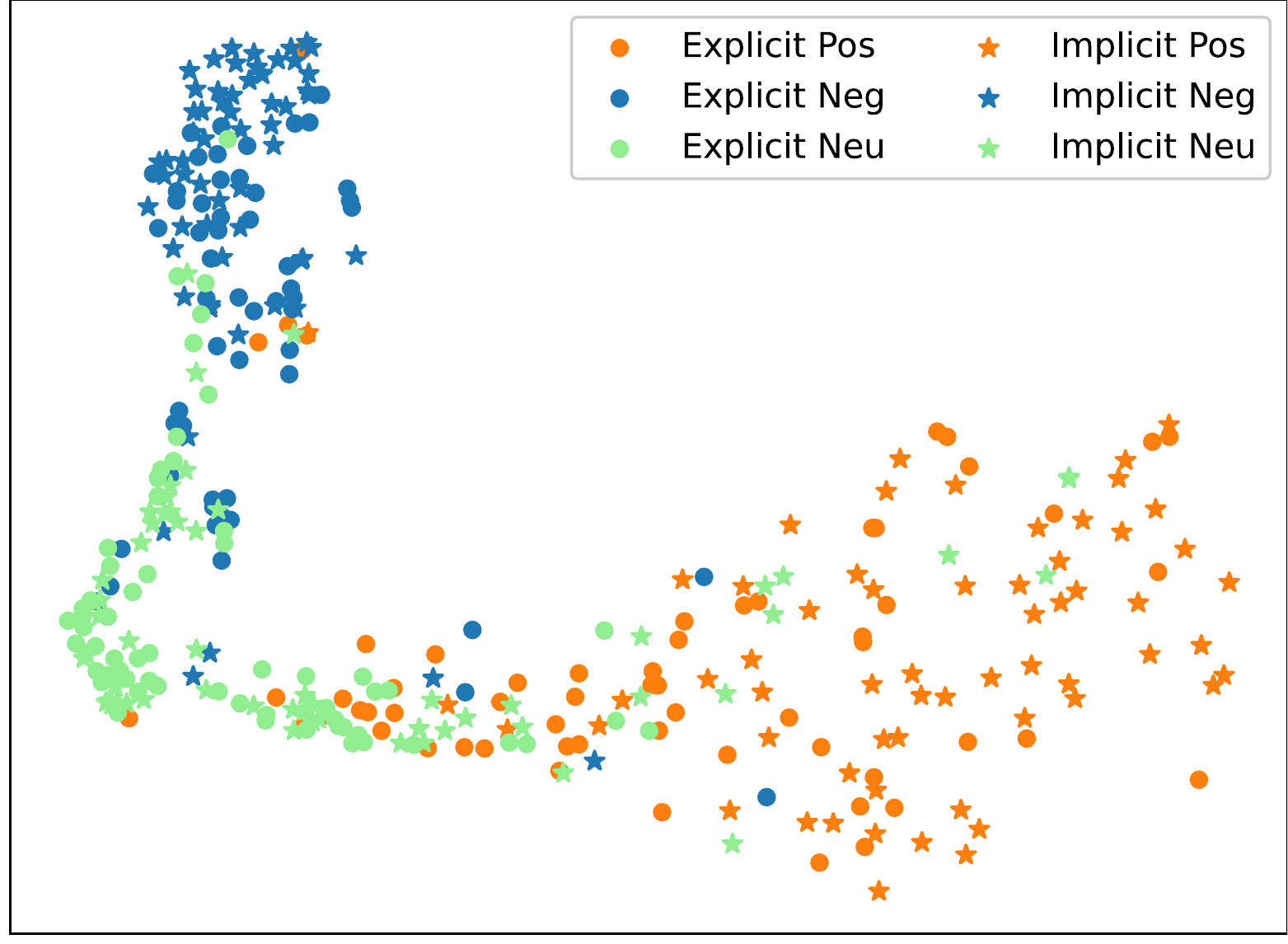}
  }
  \vspace{-2mm}
  \caption{\label{fig: cluster} Visualization of the hidden sentiment representations on Restaurant (best to view the colored version). BERTAsp+SCAPT tightly clusters the representations of both explicit and implicit sentiment expressions.}
\end{figure*}

\renewcommand\tabcolsep{3mm}
\newcolumntype{X}{C{0.07\textwidth}}

\begin{table}[t]
\small
\centering
\begin{spacing}{1.19}
\begin{tabular}{l|XX}
\bottomrule
\multirow{2}{*}{\textbf{Method}} & \multicolumn{2}{c}{\textbf{Restaurant}} \\
\cline{2-3}
 & Acc. & F1 \\
\hline
TransEncAsp & 77.10 & 57.92 \\
TransEncAsp+SCAPT & 83.39 & 74.53 \\
BERTAsp & 85.71 & 78.95 \\
BERTAsp+SCAPT & 89.11 & 83.79 \\
\ \ (-SCL) & 86.73 & 80.90 \\
\ \ (-MAP) & 88.13 & 83.22 \\
\ \ (-RR) & 87.95 & 82.38 \\
\ \ (-MAP-RR) & 87.14 & 81.10 \\
\toprule
\end{tabular}
\end{spacing}
\caption{Ablation study on SCAPT.}
\label{tab:ablation}
\end{table}

\subsection{Implicit Sentiment Learning in SCAPT}
We conclude the key aspects of learning implicit sentiment in SCAPT as exposing to sentiment knowledge and using supervised contrastive learning. 
The results in Table \ref{tab:main} shows that implicit sentiment is more challenging to learn than explicit sentiment, and previous methods based on attention or syntax modeling are not tackling the issue perfectly.
The knowledge-enhanced baselines perform slightly better with ~5\% performance gain on ISE.
By pre-training on large-scale sentiment-annotated corpora, our models achieve remarkable performance improvement on implicit sentiment learning, with 19.59\%/29.62\% relative gain on TransEncAsp.
These results prove that in-domain sentiment knowledge is absolutely necessary for implicit sentiment learning, which is provided by our retrieved corpora.
Furthermore, the models pre-trained with supervised contrastive learning objective surpasses cross-entropy classification in ISE slices.
Compared with BERTAsp+CEPT, BERTAsp+SCAPT is 4.49\%/1.73\% better on ISE, which leads to its better performance on the whole tasks. The deployment of supervised contrastive learning objective enhances noise immunity of the pre-training process, thus the pre-trained models are more effective in learning implicit sentiment.

\subsection{Ablation Study on SCAPT}

As illustrated in Table \ref{tab:ablation}, we validate the effectiveness of each part by ablation study. 
First, removing supervised contrastive learning loss (-SCL) leads to a 2.38\% performance drop on Restaurant, which is more significant than the occation of removing the other two objectives (-MAP-RR). 
This verifies that supervised contrastive learning plays a primary role in SCAPT. 
Besides, we observe that the removing of masked aspect prediction and review reconstruction objectives also brings about performance drop. This demonstrates that these mechanisms are also indispensable in SCAPT.

\subsection{Hidden Sentiment Representations}
For better understanding the behavior of our proposed methods, we further perform a visualization of the sentiment representation using t-SNE \cite{van2008visualizing}. 
As seen in Figure \ref{fig: cluster}, models with sentiment pre-training have a strong embedding ability for sentiment expression, while many misclassifications can be found in BERTAsp.
The visualization also shows that BERTAsp+SCAPT tightly clusters the representations of both implicit and explicit sentiment expressions.

\subsection{Aspect Robustness}
We analyze the robustness of our proposed models on aspect robustness test sets. Aspect robustness of ABSA was first emphasized and tested in \citet{xing-etal-2020-tasty} by applying several perturbations on reviews from Restaurant and Laptop. TextFlint \citep{wang-etal-2021-textflint} extended these transformations by introducing transformations from various linguistic perspectives. The test sets are designed to probe whether models could distinguish the sentiment of the target aspect from the non-target aspects and unrelated information.

Table \ref{tab:robust} lists the performance of tested models, in which the robustness of our proposed models is convincingly proved.
Comparing to obvious performance drop in baseline models, BERTAsp+SCAPT performs significantly better than other models with 9.05\%/6.63\% decline on Restaurant and Laptop. The results show that models pre-trained with SCAPT are more robust for aspect-level perturbations, which attribute to the better modeling for sentiment and context information with the enhancement of in-domain sentiment knowledge.

\renewcommand\tabcolsep{3.5pt}
\begin{table}[t]
\scriptsize
\centering
\begin{spacing}{1.19}
\begin{tabular}{l|cc|cc}
\bottomrule
\multirow{2}{*}{\textbf{Method}} & \multicolumn{2}{c|}{\textbf{Restaurant-test}} & \multicolumn{2}{c}{\textbf{Laptop-test}} \\
 & Ori $\to$ New & Decline & Ori $\to$ New & Decline \\
\hline
LSTM & 75.98$\to$14.64 & -61.34 & 67.55$\to$9.87 & -57.68 \\
ASGCN & 77.86$\to$24.73 & -53.13 & 72.41$\to$19.91 & -52.50 \\
CapsNet+BERT & 83.48$\to$55.36 & -28.12 & 77.12$\to$25.86 & -51.46 \\
BERT & 83.04$\to$54.82 & -29.22 & 77.59$\to$50.94 & -26.65 \\
BERT-PT & 86.70$\to$59.29 & -27.41 & 78.53$\to$53.29 & -25.24 \\
\hline
TransEncAsp+SCAPT & 83.39$\to$67.76 & -15.63 & 76.80$\to$52.52 & -24.28 \\
BERTAsp+SCAPT & \textbf{89.11$\to$80.06} & \textbf{-9.05} & \textbf{82.76$\to$76.13} & \textbf{-6.63} \\
\toprule
\end{tabular}
\end{spacing}
\caption{Model performance on aspect robustness test sets. We compare the model accuracy on the original and new test sets, and the decline of prediction on new examples are reported.}
\label{tab:robust}
 \end{table}

\section{Related Work}
\paragraph{Neural Network Methods for ABSA}
The early neural network methods \citep{wang-etal-2016-attention, ijcai2017-568} in ABSA employed various of attention mechanisms to identify aspect-related context. 
Memory Network \citep{tang-etal-2016-aspect, chen-etal-2017-recurrent-attention, wang-etal-2018-target} was further proposed to identify corresponding sentiment expression for aspects.
Recent efforts \citep{he-etal-2018-effective, tang-etal-2020-dependency} used syntax information from dependency trees to enhance attention-based models. 
A lot of works \citep{zhang-etal-2019-aspect, sun-etal-2019-aspect, wang-etal-2020-relational} make use of graph neural networks to incorporate tree-structured syntactic information and capture aspect-related information in text. 
Another line in ABSA concentrated on utilizing external corpus and pre-trained knowledge to enhance semantic awareness of models \citep{xu-etal-2019-bert, rietzler-etal-2020-adapt, dai-etal-2021-syntax}. 

\paragraph{Contrastive Representation Learning}

Our work adopts contrastive method in representation learning to acquire discriminating instance representations. 
Recent work on contrastive representation learning of instances usually based on estimating representation similarities on similar and dissimilar pairs, which are usually composed in a self-supervised manner \citep{chen2020simple, he2020momentum}. 
Specially, \citet{khosla2020supervised} illustrated a supervised contrastive method to build positive pairs between instances with same class label, and put their representations together. 
In this work, our models learn to capture implicit sentiment from informative but noisy language resources in supervised contrastive pre-training. 

\section{Conclusion}

In this paper, we introduce Supervised ContrAstive Pre-Training (SCAPT) for ABSA. By noticing that implicit sentiment is not well-handled by current neural network based ABSA models, we argue that more sentiment knowledge 
is required to solve this issue. We therefore retrieve large-scale in-domain 
annotated corpora, and propose SCAPT to learn sentiment knowledge from the corpora. 
Experimental results show that our proposed models with SCAPT achieve SOTA performance. Moreover, SCAPT is proven to be effective in implicit sentiment learning.
We hope to inspire future researches on learning and modeling implicit sentiment with knowledge-enhanced methods.

\section*{Acknowledgments}
The authors wish to thank the anonymous reviewers for their helpful comments. This work was partially funded by China National Key R\&D Program (No.\,2017YFB1002104), National Natural Science Foundation of China (No.\,62076069, 61976056), Shanghai Municipal Science and Technology Major Project (No.\,2021SHZDZX0103).


\bibliography{anthology,custom}
\bibliographystyle{acl_natbib}





\end{document}